\documentclass[letterpaper, 10 pt, conference]{ieeeconf}  
\usepackage{amsmath}
\usepackage{graphicx} 
\usepackage[ruled,vlined]{algorithm2e}
\usepackage{cite}
\usepackage{multirow}
\usepackage{caption} 
\usepackage{subcaption}

\graphicspath{{./figs/}}
\IEEEoverridecommandlockouts                              

\title{\LARGE \bf
  A Deep Reinforcement Learning Approach for \\Fair Traffic Signal Control
}

\author{Majid Raeis and Alberto Leon-Garcia 
\thanks{The authors are with the Department of Electrical and Computer Engineering, University of Toronto, ON M5S 1A1, Canada \hspace{2cm}
{ \tt\small  \noindent \{m.raeis, alberto.leongarcia\}@utoronto.ca}}
}

\begin{document}

\maketitle
\thispagestyle{empty}
\pagestyle{empty}

\begin{abstract}
Traffic signal control is one of the most effective methods of traffic management in urban areas. In recent years, traffic control methods based on deep reinforcement learning (DRL) have gained attention due to their ability to exploit real-time traffic data, which is often poorly used by the traditional hand-crafted methods. While most recent DRL-based methods have focused on maximizing the throughput or minimizing the average travel time of the vehicles, the fairness of the traffic signal controllers has often been neglected. This is particularly important as neglecting fairness can lead to situations where some vehicles experience extreme waiting times, or where the throughput of a particular traffic flow is highly impacted by the fluctuations of another conflicting  flow at the intersection. In order to address these issues, we introduce two notions of fairness: delay-based and throughput-based fairness, which correspond to the two issues mentioned above. Furthermore, we propose two DRL-based traffic signal control methods for implementing these fairness notions, that can achieve a high throughput as well. We evaluate the performance of our proposed methods using three traffic arrival distributions, and find that our methods outperform the baselines in the tested scenarios.

\end{abstract}

\section{INTRODUCTION}
Traffic signal controllers play an important role in mitigating traffic congestion in urban areas, which can have significant economic and environmental impacts as a consequence. Traditionally, traffic signal controllers use fixed hand-crafted control rules, which do not consider the current condition of the traffic (e.g. fixed cycle-based methods) or have low flexibility in response to different traffic situations~\cite{webster, sotl}. As a result, various control methods have been proposed in the literature to improve the efficiency of the traffic signal controllers. 
The new advances in the field of deep reinforcement learning (DRL) and the availability of rich traffic data have led to a new line of research, where the control algorithm is dynamically learnt from iterative interactions between the traffic signal controller and the road network environment. While recent DRL-based methods have made a promising progress, most of these methods aim at maximizing the throughput or minimizing the average travel time. However, this can lead to unfair situations where some vehicles or particular traffic flows with low traffic rates can experience large delays. The trade-off between throughput and fairness is an intrinsic characteristic of the scheduling problem. More specifically, the traffic signal control is essentially a scheduling problem, where the intersection can be modeled as a server that needs to be scheduled between different traffic flows (e.g. north-south or west-east traffic flows). While fairness is an important performance measure of the scheduling algorithms, it has often been neglected in the design of the traffic signal controller. The goal of this paper is to study the notion of fairness in traffic signal control, as well as to propose two DRL-based algorithms for providing fair control of an intersection. The contributions of this paper are summarized as follows
\begin{itemize}
    \item We introduce two notions of fairness in the traffic control context, which we will refer to as \emph{delay-based fairness} and \emph{throughput-based fairness}. As the names suggest, the former metric is concerned with the  experienced delay of the vehicles, while the latter notion measures the fairness of a signal control method in terms of the achieved throughput for different traffic flows.  
    \item In order to achieve these fairness notions, we propose two DRL-based algorithms. It should be mentioned that the reward function design is the most important and challenging part of the problem, and the provided insights can be used in other DRL-based methods irrespective of the used learning algorithm.
    \item Our proposed methods also consider the system throughput, where the balance between throughput and fairness can be adjusted using a trade-off hyper-parameter. 
\end{itemize}
\subsection{Related Work}
The classical traffic signal control methods are based on hand-crafted rules that often have limited flexibility in response to real-time fluctuations of the traffic~\cite{webster, sotl}. The Webster method~\cite{webster} and the Self-Organizing Traffic Light control (SOTL)~\cite{sotl} are two important examples in this category, where the former is used for calculating the cycle length and the phase split of the traffic light, while the latter is a vehicle actuated control method. Max-Pressure~\cite{max_pressure} is a state-of-the-art control method in this category, which maximizes the throughput of the whole network under some specific conditions. 
Defining the pressure of a phase as the difference between the total queue lengths of the incoming and outgoing approaches, max-pressure picks the phase with the maximum pressure.

With the new advances of DRL and the availability of rich traffic data, the focus of the literature on traffic signal control has shifted to data-driven methods. RL-based methods use various features of the intersection's traffic as their observation, and take actions in response to the current traffic condition. Particularly, queue length of the waiting vehicles~\cite{el2013,chen2020}, waiting time~\cite{wei2018, chu2019}, image representation of the intersection~\cite{wei2018, van2016}, phase duration~\cite{el2013} and speed of the vehicles~\cite{el2010, nishi2018} are some examples of the features that have been considered in the literature. Similarly, different metrics can be used in the calculation of the reward functions~\cite{survey}. However, these features should be selected carefully as the collection of some of these features can be impractical in a real-world scenario and more importantly, superfluous information can lead to large state spaces. 

The goal that is commonly used in RL-based traffic signal control methods is to minimize the average travel time of the vehicles~\cite{survey}. However, it is often hard to optimize the travel time directly, as it is a long-term metric that depends on a sequence of actions and therefore, the impact of a single action cannot be assessed easily. As a result, the common approach is to consider other short-term metrics such as the queue length or the delay of the vehicles. In particular, the reward function is often designed as a weighted sum of various short-term traffic metrics~\cite{wei2018, el2010, el2013, van2016}.
Fairness is another important performance measure, which has often been neglected in the existing literature. In this paper, we study this less investigated topic in traffic signal control.

\subsection{RL Background}
A reinforcement learning problem consists of two major elements, the agent and the environment, which have iterative interactions with each other at each time step $t\in \{0, 1, 2,\cdots\}$. The environment is modeled by a Markov Decision Process (MDP), which is represented by $<\mathcal{S},\mathcal{A},\mathcal{P},\mathcal{R}>$, where $\mathcal{S}$ denotes the state space, $\mathcal{A}$ is the action space, $\mathcal{P}$ is the state transition probability matrix and $\mathcal{R}$ is the reward function. At time step $t$, the agent is in state $s_t\in\mathcal{S}$, takes action $a_t \in \mathcal{A}$, transits to state $s_{t+1}$ and receives a reward of $r_t=\mathcal{R}(s_t, a_t, s_{t+1})$. The agent's actions are governed by its policy, represented by $\pi(a|s)$, which is the probability of taking action $a$ in state $s$. The total discounted reward from time step $t$ onwards is called the return, which is represented by $G_t=\sum_{k=t}^\infty \gamma^{k-t} r_k$, where $0 < \gamma < 1$. Moreover, the Q-function is defined as $Q_\pi(s,a)=\text{E}_\pi [G_t|s_t=s, a_t=a]$, which denotes the expected return starting from state s, taking action a, and following policy $\pi$. The goal of the agent is to find the optimal policy $\pi^*$ that maximizes the Q-function.

\subsubsection*{Double Deep Q-Network (DDQN)~\cite{ddqn}} While a natural way to represent the Q-function is using a table where each element corresponds to the Q-value of a particular state and action, maintaining this table becomes intractable as the state/action spaces grow. In this case, we can approximate the Q-function using a neural network with parameters $\theta$, which we represent by $Q_\theta(s,a)$. Furthermore, we use a technique known as experience replay, where the experiences are stored in the replay buffer after each interaction, while a mini-batch that is randomly sampled from the buffer is used to update the network. The loss function is defined as $\mathcal{L} (\theta) = \mathbf{E}_{\pi} \{ (y_t^{DDQN} - Q_\theta (s,a))^2\}$, 
where 
\begin{equation}
\label{DDQN}
    y_t^{DDQN}  = r_{t+1} + \gamma Q_{\theta'}(s_{t+1},\text{argmax}_a Q_{\theta}(s_{t+1},a)).
\end{equation}
In the above expression, $\theta'$ corresponds to the parameters of the target network, which are updated as $\theta' \leftarrow \tau \theta + (1-\tau) \theta'$,~$\tau \ll 1$, with the goal of stabilizing the learning.

\section{System Model and Fairness Notions} \label{sec:sys_model}
We consider an intersection of two roads, where an intelligent traffic light agent controls the flow of the vehicles. We use `N', `E', `S' and `W' to represent the incoming approaches from North, East, South and West directions, respectively, and denote the set of all incoming approaches by $\mathcal{K}$, i.e. $\mathcal{K}=\{N, E, S, W\}$. Let us denote by NS and WE, the traffic flows with North-South/South-North and West-East/East-West directions, respectively. Furthermore, we define the throughput of a traffic flow at time step $t$ as the number of vehicles that pass the intersection in the flow's direction at that time step, which is denoted by $T_{NS}(t)$ ($T_{WE}(t)$) for the NS (WE) traffic flow. The traffic light has only two phases: Green-WE and Green-NS, which denote the green light for the WE and NS roads, respectively. We assume that a phase change (e.g. Green-WE to Green-NS) includes a 3 seconds yellow light, while the other direction is still kept red. Furthermore, the traffic light agent is not allowed to change the current phase if the time spent in this phase is less than $\delta_{\text{switch}}$ seconds (including the 3 seconds required for the yellow light). This is to avoid frequent phase changes, as well as providing enough time for the pedestrians to cross the road. The queue lengths of the waiting vehicles behind the traffic light are represented by $(q_{N}, q_E, q_S, q_W)$. Finally, we denote the set of all waiting vehicles at the intersection at time step $t$ by $\mathcal{N}_t$. 

The goal of our traffic light controller is to provide a balance between the system's throughput and the fairness. We consider two notions of fairness, which we refer to as \emph{delay-based fairness} and \emph{throughput-based fairness} throughout the paper. In the following, we describe these notions in more detail.
\subsubsection*{Delay-based Fairness} This notion considers the fairness from an individual vehicle's perspective. More specifically, the delay-based fairness is concerned with decreasing the number of vehicles that experience long waiting times, irrespective of the flow to which they belong. 
Since there is no standard metric to measure the fairness in this context, we use different metrics such as the tail of the vehicles' delay distribution, 0.95 quantile and the Jain's fairness index~\cite{jain}, which is often used in the communication network context.
\subsubsection*{Throughput-based Fairness} In contrast to the previous notion where all the vehicles are treated individually, here the goal is to provide fairness among different traffic flows (directions). This is particularly important when we want to minimize the effect of conflicting traffic flows on each other (e.g. WE and NS). To achieve this goal, we extend the notion of weighted fairness from computer networking context~\cite{wfq} to the urban traffic control, where each traffic flow is guaranteed to receive service proportional to some predefined fairness weight. It should be mentioned that the traffic flows' shares can increase (proportional to their fairness weights) when some of the conflicting flows have no traffic. In this paper, we focus on  fairness among the NS and WE traffic flows. However, this can be easily generalized to scenarios with more conflicting traffic flows (e.g. left-turn traffic flow). Now, assigning fairness weights $\phi_{NS}$ and $\phi_{WE}$ to the NS and WE directions, respectively, the ideal goal is to guarantee
\begin{equation} \label{eq:ideal_fair}
    \frac{T_{NS}(t_1, t_2)}{\phi_{NS}} = \frac{T_{WE}(t_1, t_2)}{\phi_{WE}},
\end{equation}
where $(t_1, t_2)$ is an arbitrary interval during which both NS and WE roads have incoming traffic. Unfortunately, the ideal goal mentioned above cannot be achieved, as the vehicles from conflicting flows cannot be served simultaneously. In the computer networking context, where the packets can be served individually, one can aim at providing worst-case upper-bounds on the violation of Eq.(\ref{eq:ideal_fair})~\cite{scfq, dscfq}. However, in traffic control context, where we do not generally assume control over individual vehicles, and the phases need to be kept fixed for a minimum duration of $\delta_{\text{switch}}$, a practical goal can be stated as minimizing the violation of Eq.(\ref{eq:ideal_fair}), i.e.,
\begin{equation}
    \min \left|\frac{T_{NS}(t_1, t_2)}{\phi_{NS}} - \frac{T_{WE}(t_1, t_2)}{\phi_{WE}}\right|.
\end{equation}
It should be noted that this goal will also improve the delay-based fairness, as it prevents starvation of a given traffic flow and therefore, reduces the probability of extreme vehicle delays.
\section{Methodology}
\begin{figure}[t!]
\centering
\includegraphics[scale=0.5]{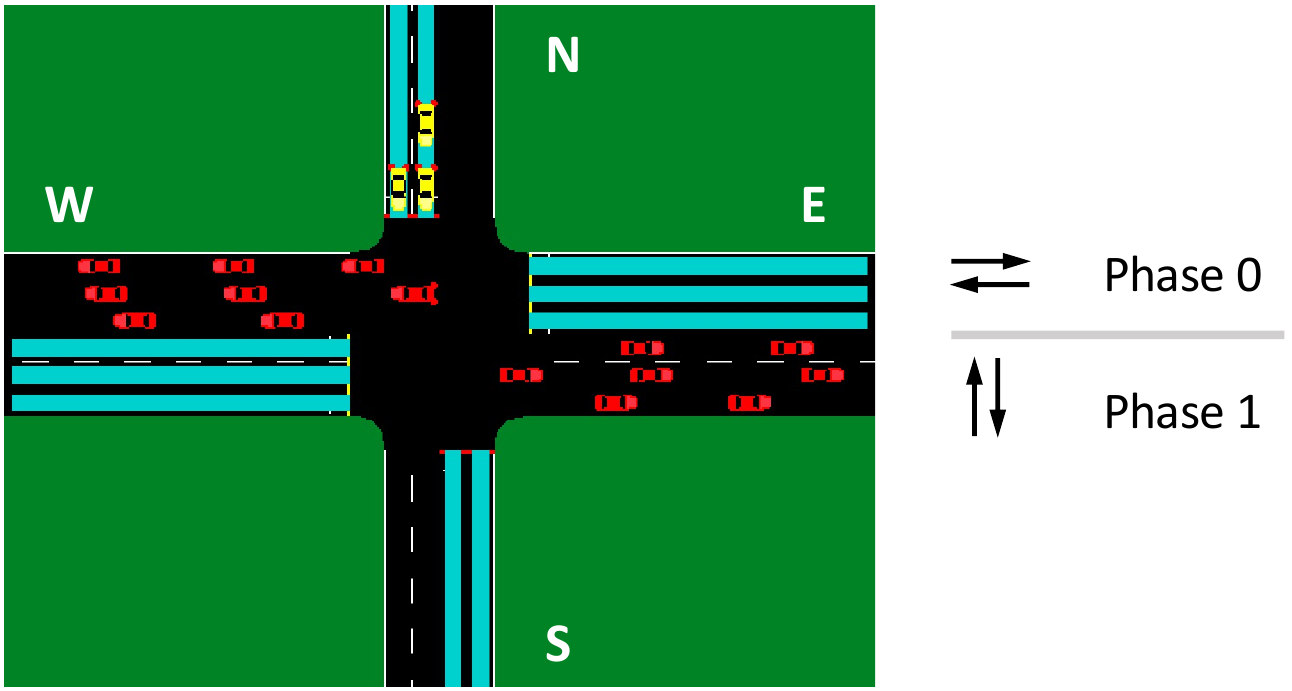}
\caption{Environment for our RL problem}
\label{fig:env}
\end{figure}

In this section, we present our method for providing fairness using a reinforcement learning approach. The environment is an intersection as shown in Fig.~\ref{fig:env}. The agent is an intelligent traffic light controller with the goal of providing delay-based or throughput-based fairness notions. In the following, we discuss the components of the RL framework, as well as the reward design process for the two introduced notions of fairness.

\subsection{Delay-based Fairness}
\noindent\textbf{State:} The state of the system includes the queue lengths of the incoming approaches (i.e., $q=(q_N, q_E, q_S, q_W)$), the sum of the waiting times of the vehicles in these approaches, which are represented by $(D_N, D_E, D_S, D_W)$,  and the current phase of the traffic light. 
While the measurement of the vehicles' waiting times at the intersections  will be possible in the near future, an alternative method for estimating the sum of the waiting times at each incoming approach is to use the queue length time sequence of that approach. In particular, after each time step, the sum of the vehicles' waiting times for a given direction (e.g. WE) increases by the queue length of the waiting vehicles behind the red light in that direction. In order to make it more clear, consider incoming approach $k$, $k\in \mathcal{K}$. Let us represent by $t_0$ the time that the traffic light becomes red for the incoming traffic in approach $k$. Furthermore, let $\Delta q_k(t)$ represent the number of vehicles that have been added to the queue of the waiting vehicles in time step $t$, i.e., $\Delta q_k(t)=q_k(t)-q_k(t-1)$. Now, the sum of the waiting times for the vehicles in approach $k$ can be calculated as $D_k(t_0+1) = q_k(t_0+1), D_k(t_0+2)= 2q_k(t_0+1)+\Delta q_k(t_0+2), D_k(t_0+3)=3q_k(t_0+1)+2\Delta q_k(t_0+2) + \Delta q_k(t_0+2), \cdots$, until the traffic light switches to green. Based on the above observation, we can calculate the sum of the waiting times (i.e., $D_k$s) using a recursive procedure that is given in Algorithm~\ref{algo:estimation}. At each time step, this algorithm updates the sum of the waiting times for approach $k$ (i.e. $D_k, k \in \mathcal{K}$) based on its values in the previous two steps, which are denoted by $D_k^{-1}$ and $D_k^{-2}$, as well as the queue lengths of the current time step and the previous time step (i.e., $q_k$ and $q_k^{-1}$).
\\
\\
\begin{algorithm}[!t]
\SetAlgoLined
\textbf{Initialization}: \\
$D_k^{-1} \leftarrow 0, D_k^{-2} \leftarrow 0, q_k^{-1} \leftarrow  0, \forall k \in \mathcal{K}$\\
\ForEach{time step}{
    \If{ $Phase \neq PreviousPhase$}{\For{all $k \in \mathcal{K}$}{$D_k^{-1} \leftarrow 0,  D_k^{-2} \leftarrow 0,  q_k^{-1} \leftarrow 0$\\}
    } 
    \For{$k \in RedPhase$ }{
    $\Delta q_k \leftarrow q_k - q_k^{-1}$\\
    $D_k \leftarrow 2D_k^{-1} - D_k^{-2} + \Delta q_k$\\
    $q_k^{-1} \leftarrow q_k$,
    $D_k^{-1} \leftarrow D_k$,
    $D_k^{-2} \leftarrow D_k^{-1}$}
    $PreviousPhase \leftarrow Phase$

}
\caption{Estimation of the sums of waiting times from the queue length sequences}
\label{algo:estimation}
\end{algorithm}

\noindent \textbf{Action:} The action of the controller at each time step is to choose one of the phases of the traffic light, i.e., Green-WE or Green-NS. The actions are taken every $\delta$ seconds, except when a phase change occurs, where the next action will be taken after $\delta_{\text{switch}}$ seconds from the phase change.
\\
\\
\textbf{Reward:} Designing the reward function is the most challenging part of the problem. While queue length based reward functions can minimize the total waiting time of the vehicles, they can lead to intolerable long waiting times for some vehicles, as they do not consider the time each vehicle has already been waiting. Therefore, we propose a reward function based on the waiting times of the vehicles, which captures throughput and fairness at the same time. There are multiple challenges in using waiting times of the vehicles in the reward function. First, the total waiting time of a vehicle will be unknown until it passes the intersection. Moreover, even if the waiting times are known beforehand, there will be a credit assignment problem. In other words, since the waiting time of a particular vehicle can be affected by multiple actions, it would not be clear how much each action is responsible for the vehicle's waiting time and therefore, how the action should be rewarded. In the following we address these issues by defining a reward function that only uses the partial waiting time of the vehicles up to the time that the reward is calculated. Moreover, an iterative algorithm similar to Algorithm~\ref{algo:estimation} can be used to estimate the waiting times of the vehicles from the queue length time sequences. Therefore, the only information required for this method is the sequence of queue lengths.

Let us denote the (partial) waiting time of vehicle $n$ up to time $t$ by $d_n(t)$. We define the immediate reward for each action as 
\begin{equation}
    r_t = - \sum_{n\in N_t} (1+\alpha (2d_n(t)-1)),
\end{equation}
where $N_t$ is the set of vehicles in the intersection at time $t$, and $\alpha$ represents a hyper-parameter for adjusting the trade-off between the throughput and fairness. Now, let us explain the rationale behind this reward function. Since the goal of the RL agent is to maximize the expected sum of rewards it receives, we first study the contribution of each vehicle's waiting time to the cumulative reward that the agent receives. Consider vehicle $n$ that has experienced a total waiting time equal to $w_n$. Representing the time that this vehicle starts to wait behind the red light by $a_n$, and noting that by definition $d_n(a_n+i) = i, 1\le i \leq w_n$, the contribution of this vehicle to the total return can be calculated as
\begin{align}
    -&\sum_{t=a_n+1}^{a_n+w_n} (1+\alpha(2d_n(t)-1)) = \nonumber\\ -&\big((1+\alpha)+(1+3\alpha)+\cdots+(1+(2w_n-1)\alpha)\big)= \nonumber\\ -&\left(w_n+\alpha w_n^2\right).
\end{align}
Representing the set of all vehicles that pass the intersection in an episode by $\mathcal{N}$, i.e., $\mathcal{N}=\cup_t \mathcal{N}_t$, we have
\begin{align}
    \text{E}\big[\sum_{t}r_t\big] &= -\text{E}\big[\sum_{n\in \mathcal{N}}(w_n+\alpha w_n^2)\big] \nonumber \\
    &= -\text{E}\Big[ \sum_{n\in \mathcal{N}} w_n\Big] - \alpha \underbrace{\text{E}\Big[ \sum_{n\in \mathcal{N}} w_n^2\Big]}_{\text{Fairness}}.\label{eq:ave_reward}
\end{align}
As we can see in Eq.~(\ref{eq:ave_reward}), the average return includes the second moments of the waiting times, which penalize the agent heavily for having extreme waiting times. Moreover, the hyper-parameter $\alpha$ adjusts the trade-off between efficiency and the delay-based fairness. Specifically, $\alpha=0$ corresponds to the case where  the total waiting time of the vehicles is minimized, while increasing $\alpha$ amplifies the effect of the fairness term. In the rest of this paper, we refer to this controller as  Delay-based Fair Controller or DFC$_\alpha$, where $\alpha$ is the fairness hyper-parameter. In order to see why increasing $\alpha$ improves the delay-based fairness, let us discuss the effect of the second term using the Jain's fairness index~\cite{jain}. This index measures the fairness (closeness) of a set of values (in our case the waiting times) as follows
\begin{equation}
    J = \frac{(\sum_{n=1}^{N}w_n)^2}{N\sum_{n=1}^N w_n^2},\label{eq:jain}
\end{equation}
which ranges from $1/N$ (worst case) to $1$ (best case). As can be observed, the fairness term and the expected sum of delays in Eq.~\eqref{eq:ave_reward} are proportional to the denominator and nominator in Eq.~\eqref{eq:jain}. Therefore, by increasing $\alpha$, the agent will be penalized more if it results in a wide range of waiting times, which decreases the fairness index. However, it should be noted that a large Jain's fairness index is not necessarily desirable either, as it means all the vehicles should experience the same waiting times, which can jeopardise the throughput of the intersection. Therefore a balance between throughput and fairness is necessary, which can be achieved by tuning hyper-parameter $\alpha$.
\subsection{Throughput-based Fairness}
Now, let us discuss the components of the RL framework for providing throughput-based fairness.
\\
\\
\noindent \textbf{State:} The state of the system consists of three components. The first component is the number of vehicles in the incoming approaches that are in a predefined vicinity of the intersection. We denote this information by $(v_N, v_E, v_S, v_W)$, where $v_k, k \in \mathcal{K}$ represents the number of vehicles in the incoming approach with direction $k$, in $L$ meters vicinity of the intersection. The second component is a parameter representing the throughput deviation of the NS and WE flows, which is represented by $\delta(t)$ and is calculated as
\begin{equation}
    \delta(t) = \sum_{\tau=0}^{t} B(\tau) \left(\frac{T_{NS}(\tau)}{\phi_{NS}} - \frac{T_{WE}(\tau)}{\phi_{WE}}\right),
\end{equation}
where $B(t)$ equals $1$ if both WE and NS directions have incoming traffic in $L$ meters vicinity of the intersection (i.e., $v_N+v_S > 0$ and $v_W+v_E > 0$), and $0$ otherwise.   
The fairness deviation is used so that the agent can keep track of the existing unfairness between the traffic flows. We can simply calculate $\delta(t)$ in an iterative manner after each time step as follows
\begin{align}
    &\delta(0) = 0, \nonumber\\ &\delta(t) = 
\delta(t-1) + B(t) \left(\frac{T_{NS}(t)}{\phi_{NS}} - \frac{T_{WE}(t)}{\phi_{WE}}\right).
\end{align}
Finally, the third component of the state is the current phase of the traffic light. 
\\
\\
\textbf{Action:} The actions are similar to the ones described for the delay-based fairness. 
\\
\\
\textbf{Reward:} 
As mentioned in the previous section, our goal is to minimize the fairness deviation between the two conflicting traffic flows. However, the throughput of the system is another metric that needs to be considered in our reward definition. Therefore, we define the immediate reward at time step $t$ as  
\begin{equation}
    r_t= - \sum_{k\in \mathcal{K}} q_k(t) - \beta |\delta(t)|, 
\end{equation}
where $\beta$ is a hyper-parameter similar to $\alpha$ in DFC, which adjusts the trade-off between fairness and throughput. Calculating the expected sum of rewards, we have
\begin{align}
    \text{E}\big[\sum_{t}r_t\big] &= -\text{E}\big[ \sum_t \sum_{k\in \mathcal{K}} q_k(t)\big] -\beta~\text{E}\big[\sum_t |\delta(t)|\big] \nonumber \\
    &= -\text{E}\Big[\sum_{n\in \mathcal{N}} w_n \Big] -\beta~\underbrace{\text{E}\Big[\sum_t |\delta(t)|\Big]}_{\text{Fairness}},\label{eq:reward_TFC}
\end{align}
where we have used the fact that the summation of the number of waiting vehicles over time ($\sum_t \sum_{k\in \mathcal{K}} q_k(t)$) equals the summation of the waiting times of all the vehicles ($\sum_{n\in \mathcal{N}} w_n$). Now, since the throughput-based fairness term in Eq.~\eqref{eq:reward_TFC} is the expected cumulative throughput deviation, increasing $\beta$ penalizes the agent more when the throughput deviations occur, while decreasing it improves the total throughput of the intersection. We will refer to this method as Throughput-based Fair Controller or TFC.
\section{Experimental Results}
In this section, we present our results on the proposed controllers' performance in terms of providing fairness. We first explain the experimental setup, baselines and traffic patterns, and then show our evaluations and results. 
\subsection{Experimental Setup}
Our experiments are conducted on the urban traffic simulator SUMO\footnote{https://www.eclipse.org/sumo/}. The environment is a four-way intersection, where the west/east road segments (major roads) are each 250 meters long and have 3 incoming and 3 outgoing lanes, while the north/south segments are 200 meters long with 2 incoming and 2 outgoing lanes. Furthermore, the WE traffic flows have a speed limit of $50~\text{km}/\text{h}$, while the NS flows have a speed limit of $30~\text{km}/\text{h}$. As mentioned earlier in Section~\ref{sec:sys_model}, the traffic light has two phases represented by Green-WE and Green-NS. Moreover, we have a 3 seconds yellow light before each phase change, and a 7-seconds minimum green time (i.e., $\delta_{\text{switch}}=10$s). Finally, the agents take action every $5$s, once the minimum green time is passed.

\subsection{Implementation Parameters}
The DFC and TFC agents are trained using DDQN. For implementing DDQN, we use a fully connected neural network with two hidden layers, which have ReLU activation functions. The last layer of the network has no activation functions as it estimates the Q-function. The models have been trained for 400 episodes, where the maximum number of training steps per episode is set to 2000. In order to achieve an efficient learning, we might terminate an episode before reaching 2000 time steps, if the queue lengths of the roads exceed a threshold of 100 vehicles. This is particularly important in the beginning  episodes, where  random  exploration  can  easily lead to heavy congestion. In this case,  the agent gets stuck in these congested states for a long time and cannot fix the situation as it has not been trained enough.
The parameters used for training our models are summarized in Table~\ref{table:hyper_parameters}. 

\begin{table}[t]
    \centering
        \caption{DDQN hyper-parameters}
        \begin{tabular}{ l  l }
        \noalign{\hrule height 0.03cm}
            Parameter & Value \\
            \hline 
            Adam optimizer learning rate &  0.001\\
            Replay memory buffer size & 1,000,000 \\
            Mini-batch size & 32 \\
            Discount factor for target & 0.99 \\
            Size of hidden layers & (32, 32) \\
            $\epsilon$ for exploration & $0.5$ to $0.05$\\
            \noalign{\hrule height 0.05cm}
        \end{tabular}
    \label{table:hyper_parameters}
\end{table}

\begin{table}[t]
    \centering
        \caption{Details of the traffic datasets}
        \begin{tabular}{ l  l }
        \noalign{\hrule height 0.03cm}
            Parameter & Value \\
            \hline
            \multicolumn{2}{c}{Poisson}  \\
            \hline 
            Arrival rate (WE/EW) &  0.2 (veh/s)\\
            Arrival rate (NS/SN) &  0.066 (veh/s)\\
            \hline
            \multicolumn{2}{c}{MMPP (NS/SN)}  \\
            \hline 
            $p_{\text{on}\to\text{off}}$ & 0.28 \\
            $p_{\text{off}\to\text{on}}$ & 0.02 \\
            \hline
            \multicolumn{2}{c}{NHPP (NS/SN)}  \\
            \hline 
            \multirow{2}{*}{$\lambda(t)$ (periodic with period = 2000s)} & 0.25 if $t \in [0, 500)$\\
            & 0.1 if $t \in [500, 2000)$\\
            \noalign{\hrule height 0.05cm}
        \end{tabular}
    \label{table:dataset}
\end{table}

\subsection{Baselines}
We compare our algorithms with the following baselines.
\subsubsection{Self-organizing Traffic Lights (SOTL)~\cite{sotl}} This method uses a number of hand-engineered rules to pick the next phase. Specifically, the current phase of the traffic light can change to the next phase only if all of these conditions are satisfied: I) the passed time since the last phase change is greater than the predefined minimum green-time, II) the number of vehicles waiting behind the red light is greater than a threshold, and III) the number of vehicles approaching the green light is in a particular range, which needs to be tuned.
\subsubsection{Max-pressure~\cite{max_pressure}} This control method is based on the concept of pressure between the incoming and outgoing lanes. Specifically, defining the pressure of a phase as the difference between the total queue lengths of the incoming and outgoing approaches, Max-pressure chooses the phase with the maximum pressure. 

\begin{figure*}[!t] 
\centering
\subcaptionbox{ }{\includegraphics[width=.27\linewidth, height=2.9cm]{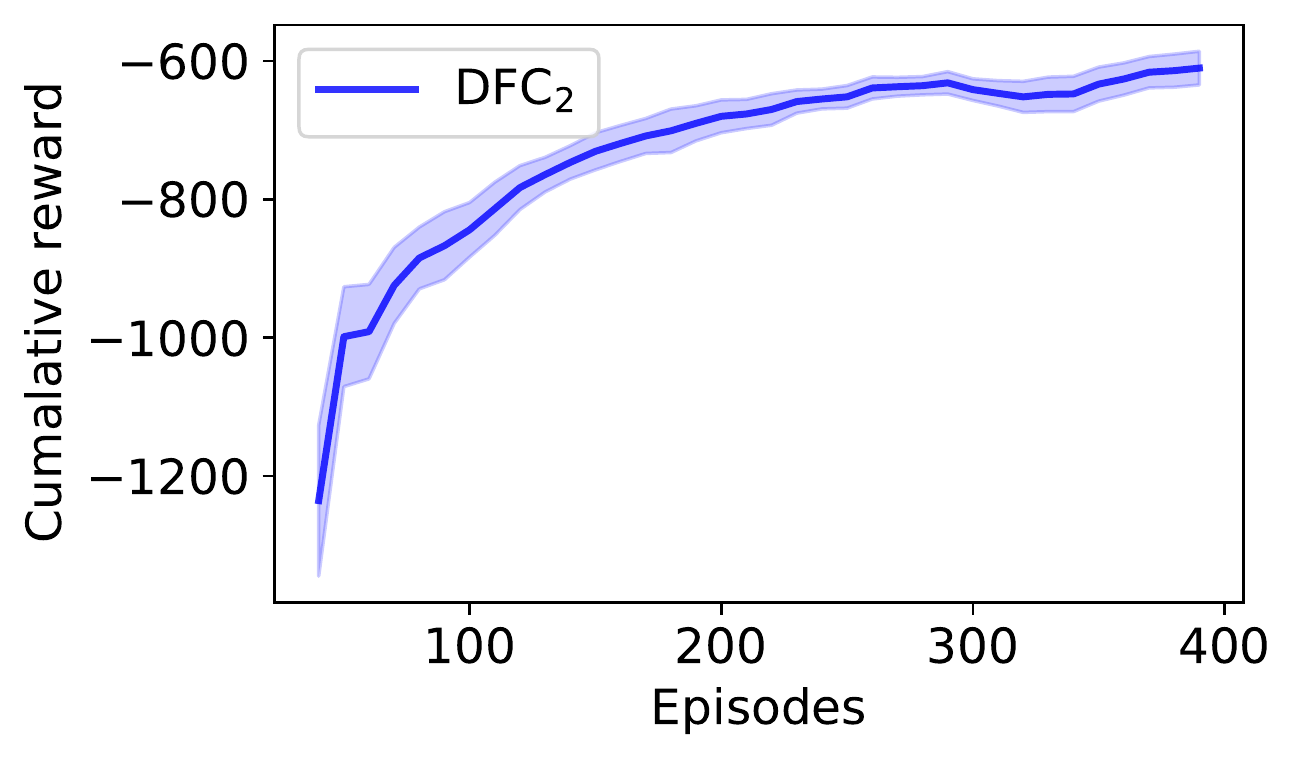}}
\subcaptionbox{}{\includegraphics[width=.27\linewidth, height=2.9cm, ]{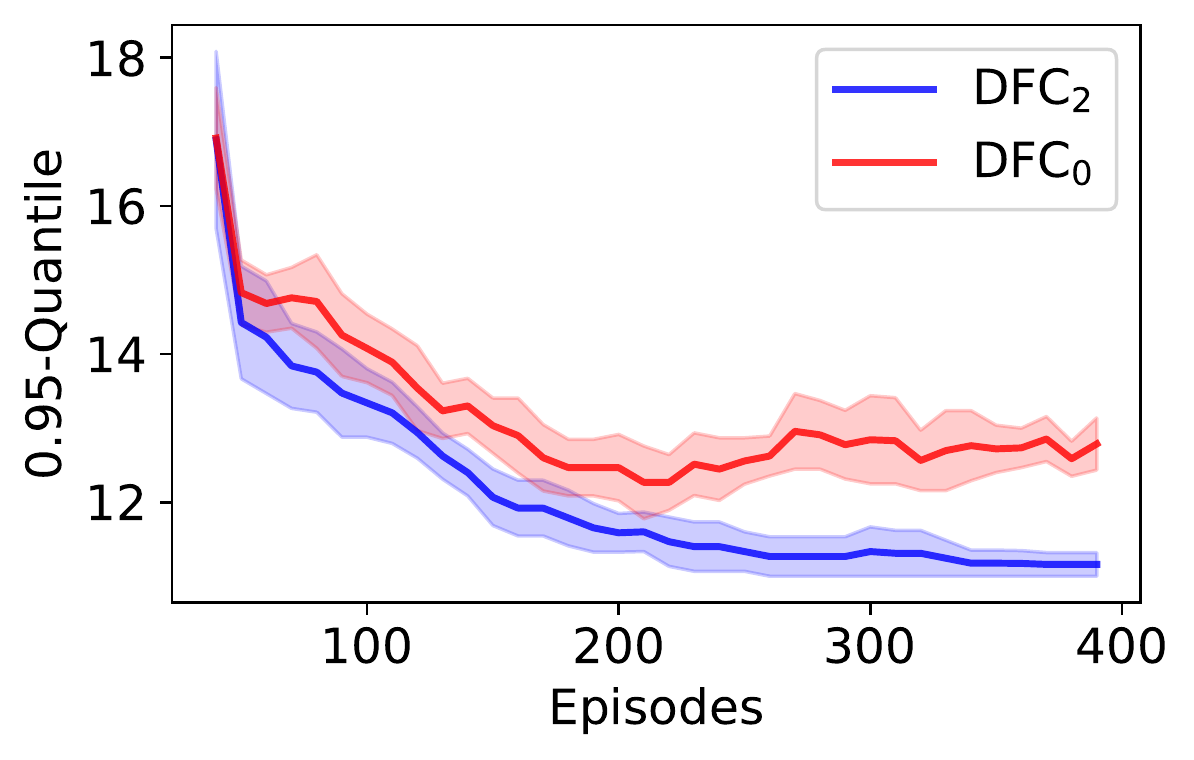}}
\subcaptionbox{}{\includegraphics[width=.27\linewidth, height=2.9cm]{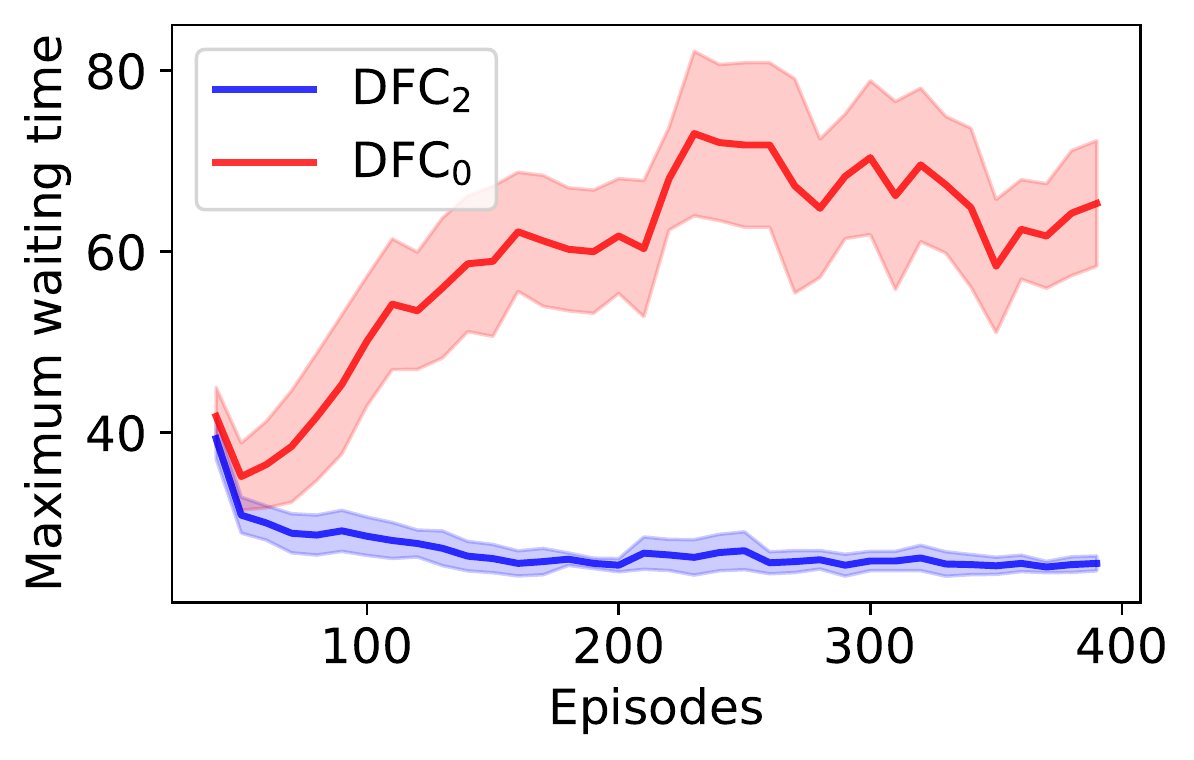}}

\caption{Learning curves: a) Cumulative reward of DFC$_2$, b and c) Comparison of the delay-based fairness between DFC$_0$ and DFC$_2$.}
\label{fig:training}
\end{figure*}
\subsection{Datasets}

  In contrast to the existing studies, which are most often limited to the Poisson arrivals, we use two other families of arrival distributions: Markov modulated Poisson process (MMPP) and non-homogeneous Poisson process (NHPP). While the Poisson process is from the renewal process family (stationary with iid inter-arrival times), MMPP and NHPP generalize the traffic arrival patterns to non-renewal (correlated inter-arrival times) and non-stationary (time-varying) processes, which are more realistic traffic models. Specifically, we use Markov modulated ON/OFF process to generate burst arrivals, which can represent platoons of vehicles. In this case, we consider two states (ON and OFF), where the Poisson traffic is generated only when the state is ON. The transition probability from state ON to OFF (OFF to ON) is represented by $p_{\text{on}\to\text{off}}$ ($p_{\text{off}\to\text{on}}$). On the other hand, NHPP is similar to the Poisson process, with the difference that the arrival rate is a function of time ($\lambda(t)$). 

\begin{figure}[!t] 
\centering
\subcaptionbox{WE}{\includegraphics[width=.49\columnwidth, height=2.6cm]{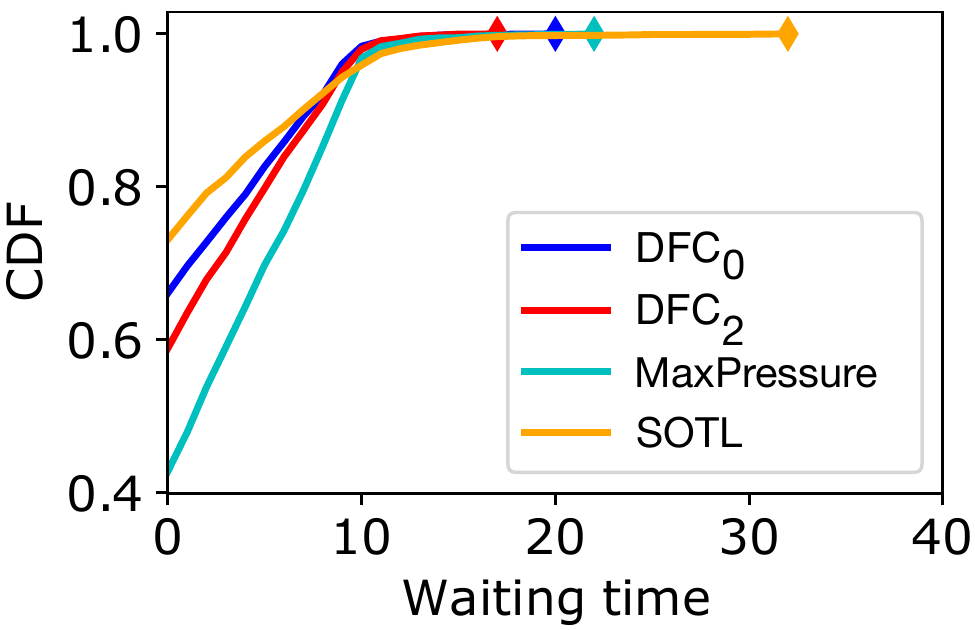}}
\subcaptionbox{NS}{\includegraphics[width=.49\columnwidth, height=2.6cm]{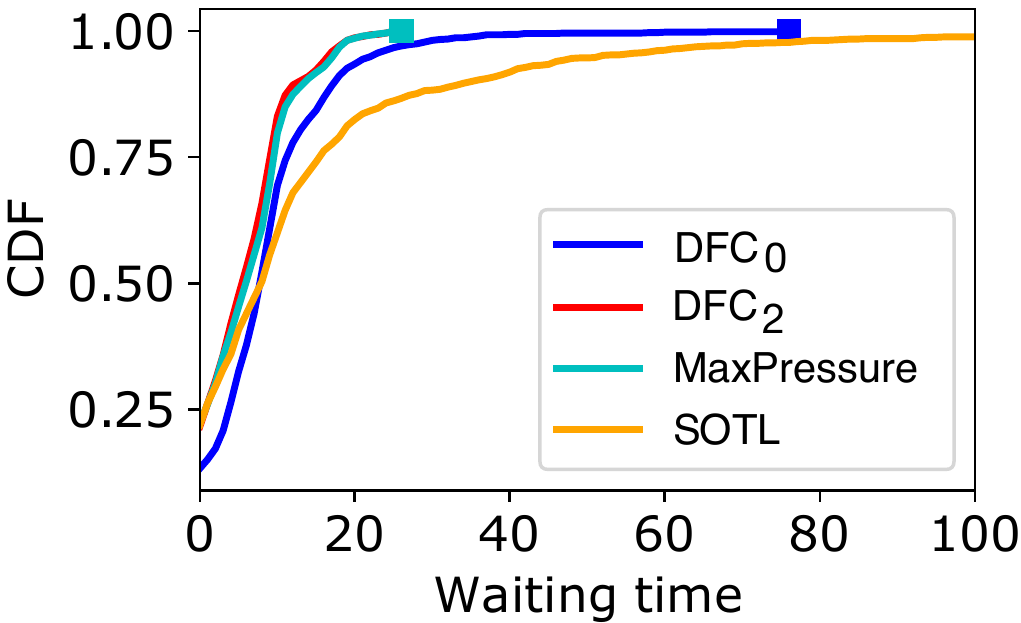}}
\caption{Cumulative distribution functions (CDFs) of the waiting times (WE: Poisson, NS: MMPP)}
\label{fig:cdfs}
\end{figure}

\subsection{Delay-based Fairness Evaluation}
\subsubsection{Training}
We first evaluate the performance of our proposed algorithms in a scenario in which the WE/EW vehicle arrival times have Poisson distribution, while the NS/SN traffic flows are generated using MMPP. The reason for this choice of traffic patterns is that the WE/EW roads represent major roads and therefore have uniform continuous patterns, while the NS/SN are minor roads which have bursty traffic patterns. It should be mentioned that our method is not based on any particular arrival distribution and these assumptions have only been made to generate the dataset for this experiment. The parameters of these traffic patterns are summarized in Table~\ref{table:dataset}.

As mentioned in the previous section, hyper-parameter $\alpha$ can be used to adjust the importance of the fairness term in the reward function. In order to validate the capability of DFC$_\alpha$, $\alpha>0$, in resulting a more fair policy, we present the results for $\alpha=0$ (no fairness) and $\alpha=2$. The convergence of the cumulative reward for $\alpha=2$ can be seen in Fig.~\ref{fig:training}a. The DFC$_2$ agent has been trained with 5 different initial seeds. The solid curve and the pale region around it show  the  average  and  the  standard  error  bands,  respectively. Since the reward functions of DFC$_0$ and DFC$_2$ are different and they can get different ranges of values, their comparison in the same figure will not be informative. To compare these two models, we use the 0.95 quantile of the waiting times, and the maximum waiting time in each episode. Figs.~\ref{fig:training} (b-c) show this comparison. We can observe that DFC$_2$ has a much better performance in terms of reducing the 0.95 quantile and the worst case waiting times.
\subsubsection{Test}
Now, let us compare the performance of our trained algorithms with the baseline methods on the test dataset.
Fig.~\ref{fig:cdfs} shows the CDFs of the waiting times for the WE and NS traffic flows separately.
The maximum waiting times are represented by the markers in the figure. For the WE traffic flow, we can observe that DFC$_2$ results in the lowest maximum waiting time. Although SOTL results in more zero waiting times for the WE traffic, the distribution of the waiting times has a longer tail for this method, which means a larger number of vehicles experience extreme waiting times. This is particularly the case for the NS traffic flow, where the distribution of the waiting times for the SOTL method has a much longer tail compared to other methods. On the other hand, DFC$_2$ and max-pressure has the best performance for the NS/SN traffic flow. Therefore, both traffic flows considered, DFC$_2$ has the best performance among the compared methods.   

\begin{figure*}[!t] 
\centering
\subcaptionbox{TFC (ours)}{\includegraphics[width=.27\linewidth]{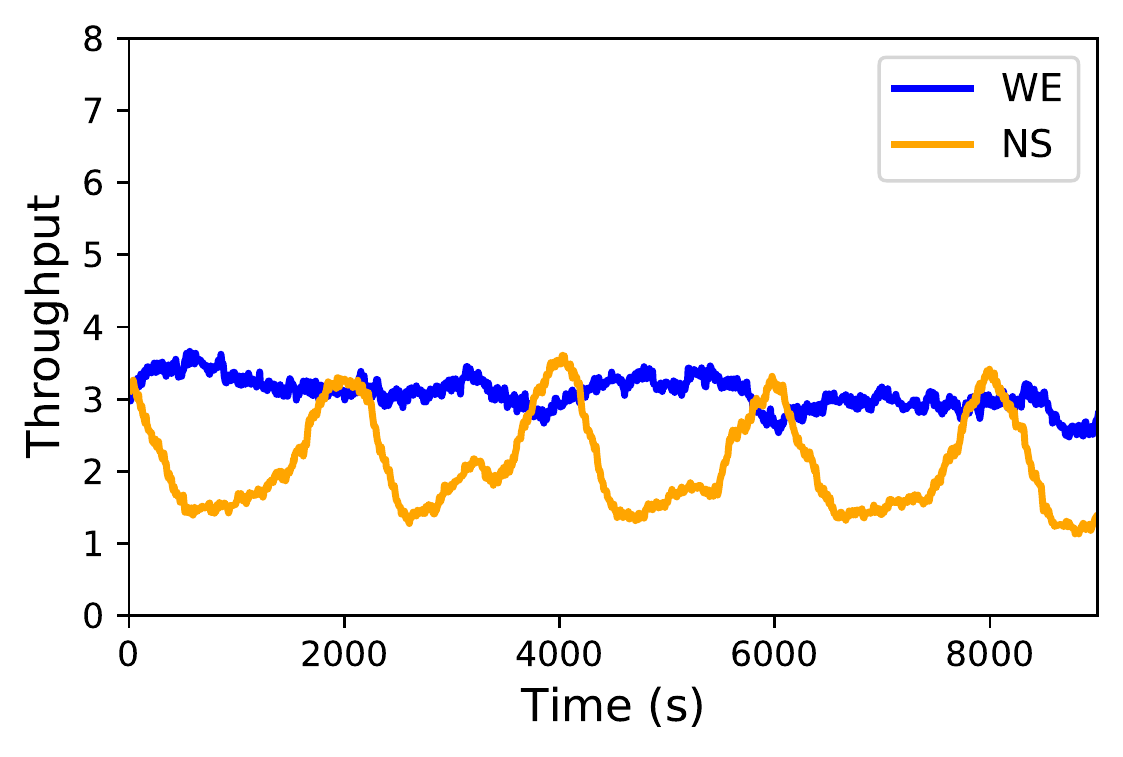}}
\subcaptionbox{Max-pressure}{\includegraphics[width=.27\linewidth]{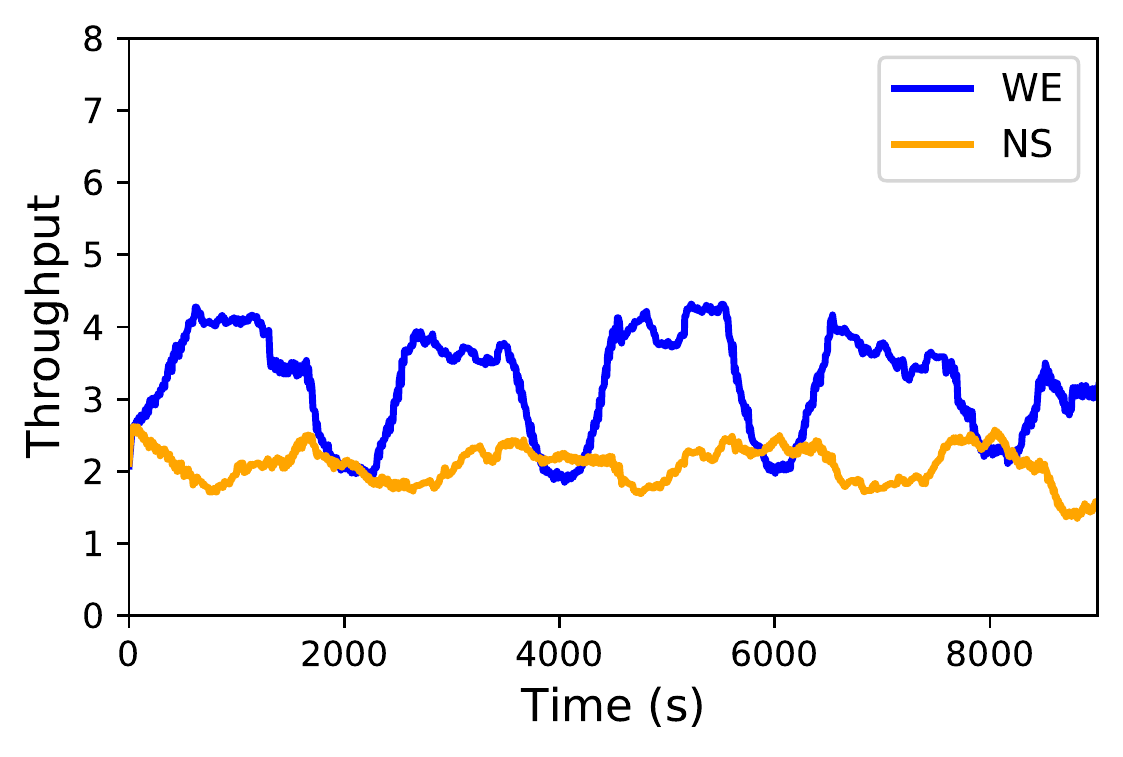}}
\subcaptionbox{SOTL}{\includegraphics[width=.27\linewidth]{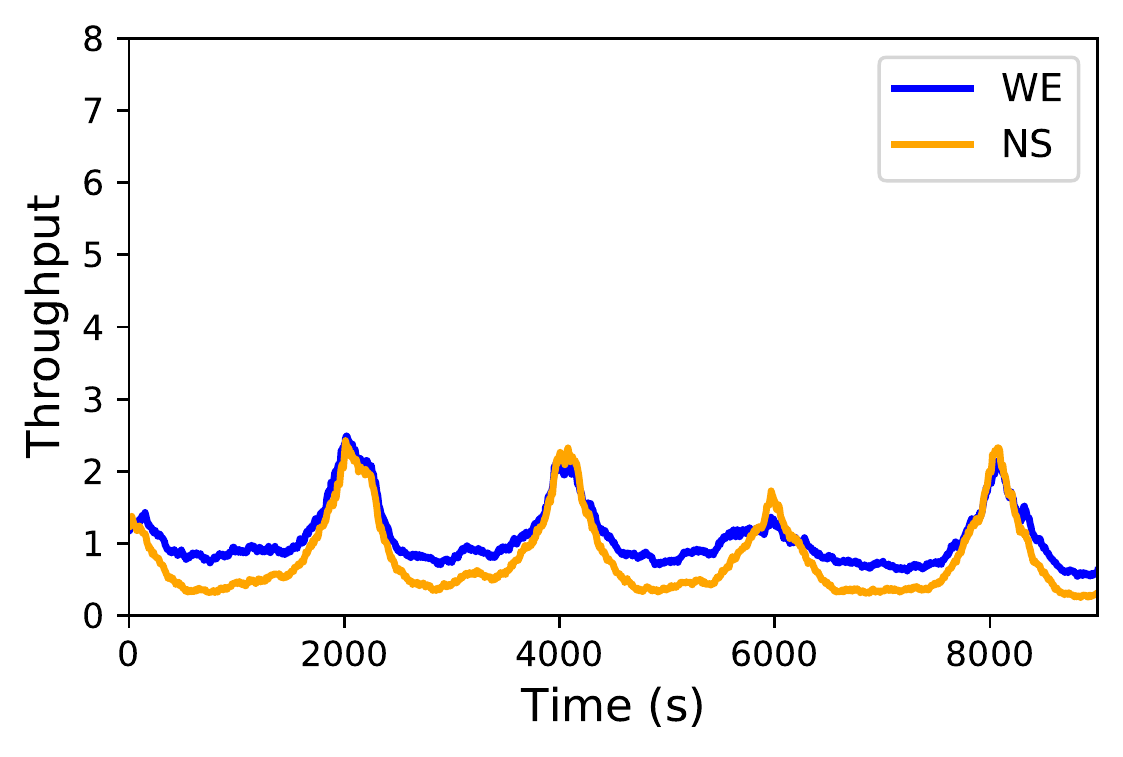}}
\hfill
\subcaptionbox{TFC (ours)}{\includegraphics[width=.27\linewidth]{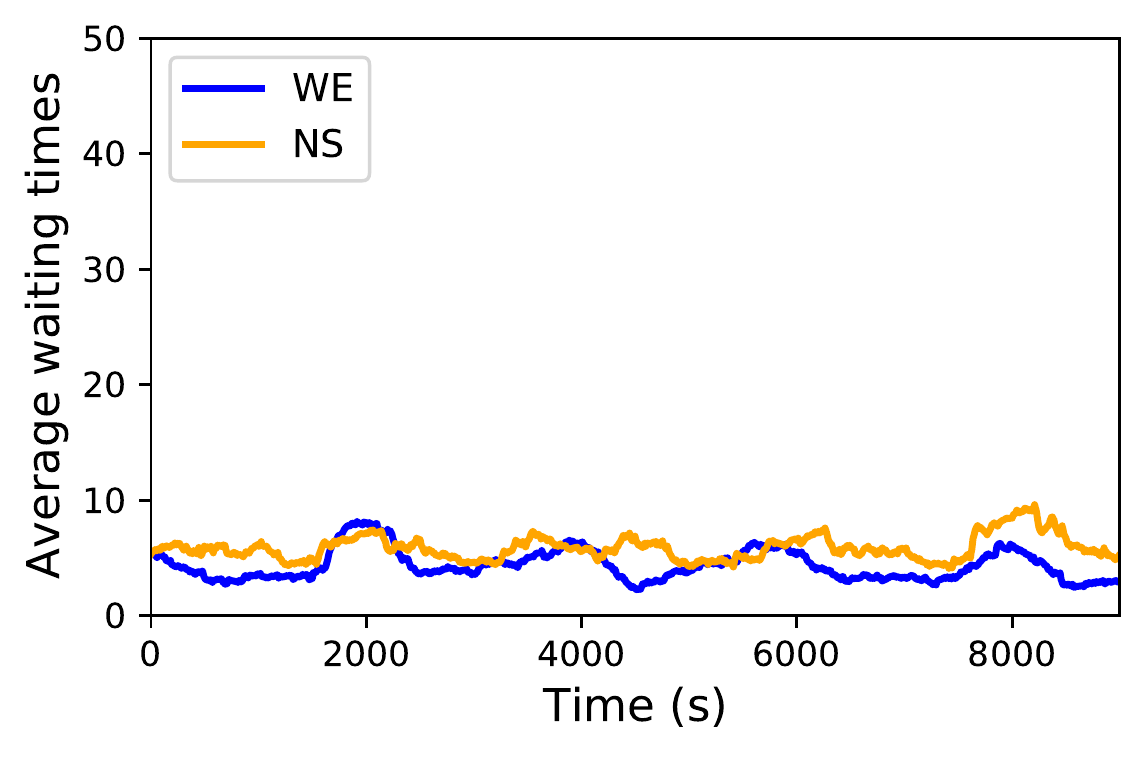}}
\subcaptionbox{Max-pressure}{\includegraphics[width=.27\linewidth]{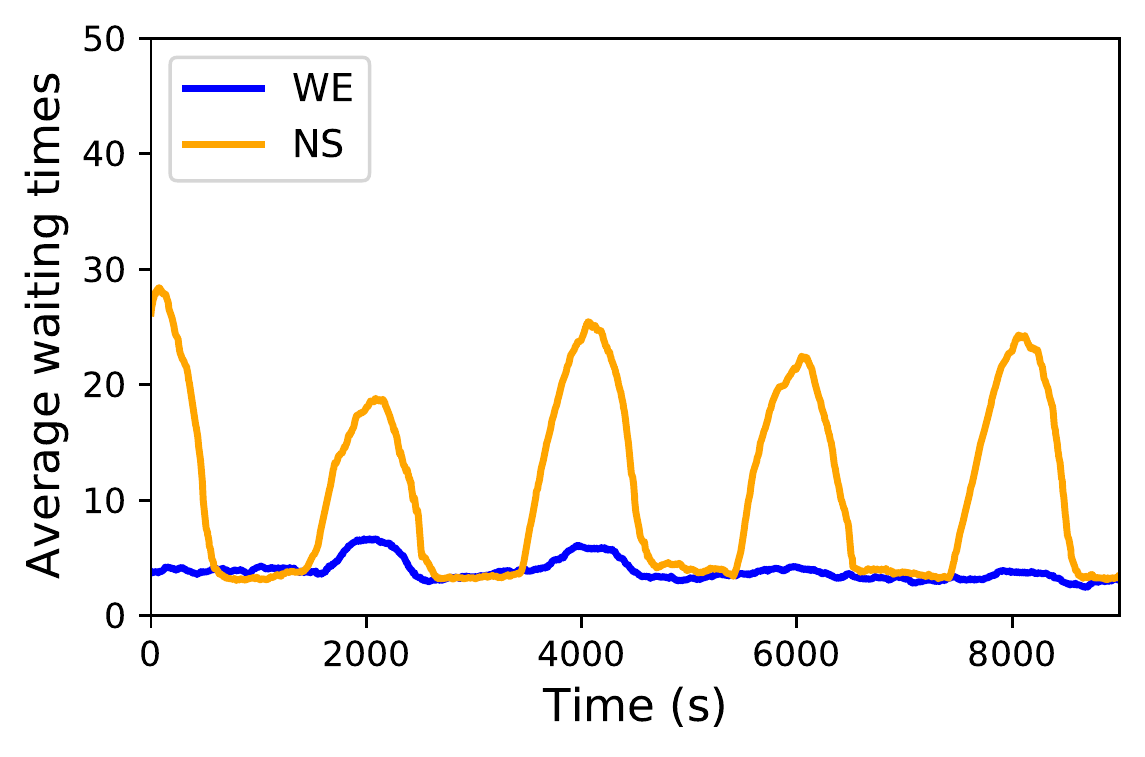}}
\subcaptionbox{SOTL}{\includegraphics[width=.27\linewidth]{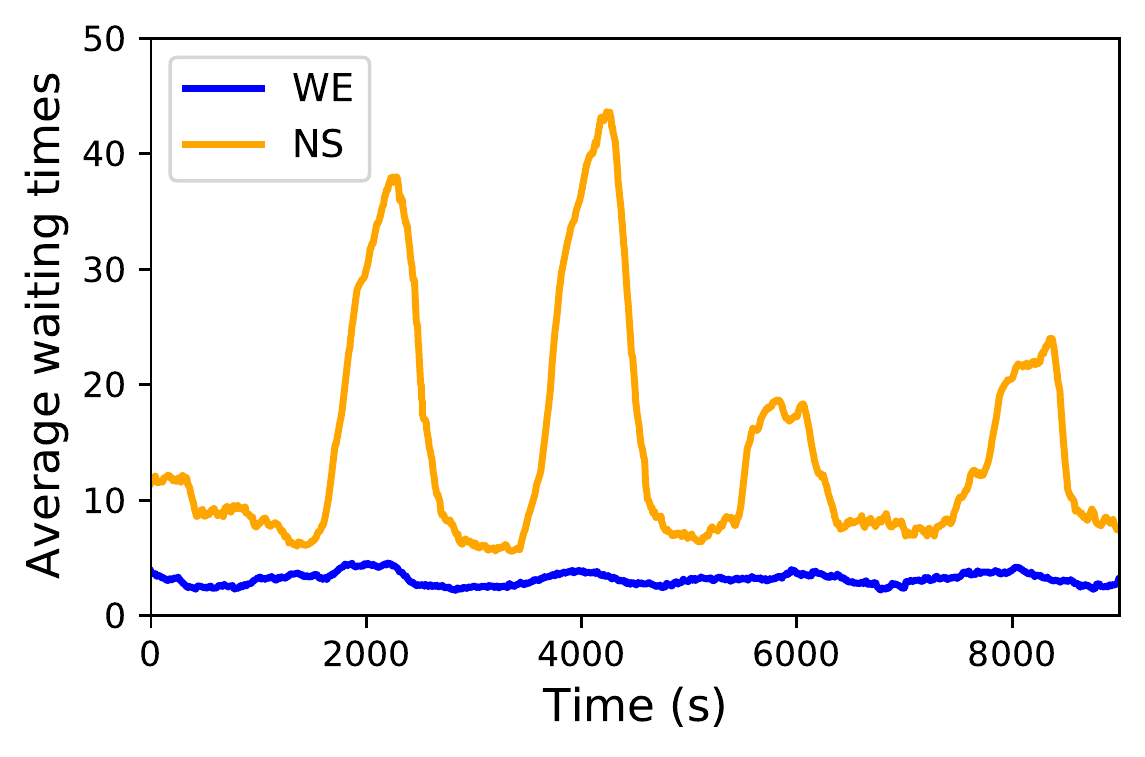}}
\caption{Comparison of the performance of TFC, Max-pressure and SOTL (WE:Poisson, NS:NHPP)}
\label{fig:nhpp}
\end{figure*}

\subsection{Throughput-based Fairness Evaluation}
In this section, we study the performance of TFC and the other baselines from the throughput-based fairness point of view.  We consider the same intersection as in the previous experiment, with the difference that the NS/SN traffic flows are time dependant and are generated by NHPP. The arrival rates for both WE and NS traffic flows are reported in Table~\ref{table:dataset}. As can be observed, there are periodic surges in the NS traffic flow, which can potentially affect the flow of traffic in the WE/EW roads. Since WE/EW are the major arterial roads, it is desirable that we provide a stable service to these traffic flows, which is not highly dependent on the variations of the NS/SN traffic. The fairness weights for the WE and NS traffic flows are set based on the average arrival rates of each direction ($\phi_{WE}=1.5$ and $\phi_{NS}=1$). The other parameters of TFC are set to $L=40$m and $\beta=0.01$.

Fig.~\ref{fig:nhpp} shows how TFC, Max-pressure and SOTL perform in this scenario. As we can observe, for Max-pressure and SOTL controllers, the throughput of the WE traffic flow is highly affected by the variations of the NS traffic. However, our controller (TFC) has been able to minimize the effect of the NS traffic flow's variations on the WE flow, which results in a stable throughput for the major road. Furthermore, Figs.~\ref{fig:nhpp}~(d-f) show that our controller has a much better performance in terms of the average waiting times (for both WE and NS flows) compared to the Max-pressure and SOTL methods, which result in large fluctuations of the waiting times for the NS traffic flow.




\section{Conclusions}
In this paper, we study the notion of fairness in traffic signal control, which is an important but less investigated problem. We introduce two notions of fairness that we refer to as delay-based and throughput-based fairness. Furthermore, we propose two DRL-based control methods for providing these fairness notions, while achieving a high throughput at the same time. We conduct experiments using three different vehicle arrival distributions, including Poisson process (renewal), Markov modulated Poisson process (non-renewal)  and non-homogeneous Poisson process (non-stationary). Our experiments validate the effectiveness of our methods, as well as their superiority over the compared baselines.
\bibliographystyle{IEEEtran}
\bibliography{reference}

\end{document}